\def\year{2021}\relax
\documentclass[letterpaper]{article} 
\usepackage{aaai21}  
\usepackage{times}  
\usepackage{helvet} 
\usepackage{courier}  
\usepackage[hyphens]{url}  
\usepackage{graphicx} 
\def\UrlFont{\rm}  
\usepackage{natbib}  
\usepackage{caption} 
\usepackage{graphicx}
\usepackage{algorithm}
\usepackage{algorithmic}
\usepackage{amsmath}
\usepackage{amsthm}
\usepackage{booktabs}

\usepackage{multirow}
\usepackage{color}
\usepackage{amsfonts}
\usepackage{bm}
\usepackage{CJKutf8}
\usepackage{array}
\newcommand{\B}[1]{\textbf{#1}}
\DeclareMathOperator*{\argmax}{arg\,max}
\newcolumntype{L}[1]{>{\raggedright\let\newline\\\arraybackslash\hspace{0pt}}m{#1}}
\newcolumntype{C}[1]{>{\centering\let\newline\\\arraybackslash\hspace{0pt}}m{#1}}
\newcolumntype{R}[1]{>{\raggedleft\let\newline\\\arraybackslash\hspace{0pt}}m{#1}}

\title{Stylized Dialogue Response Generation Using Stylized Unpaired Texts}
\author {
    Yinhe Zheng\textsuperscript{\rm 1,2}\thanks{Equal contribution} ,
    Zikai Chen\textsuperscript{\rm 1}\footnotemark[1],
    Rongsheng Zhang\textsuperscript{\rm 3},
    Shilei Huang\textsuperscript{\rm 3},
    Xiaoxi Mao\textsuperscript{\rm 3}, 
    Minlie Huang\textsuperscript{\rm 1}\thanks{ Corresponding Author: aihuang@tsinghua.edu.cn} \\
}
\affiliations {
  
    \textsuperscript{\rm 1} {Department of Computer Science and Technology, Institute for Artifical Intelligence, State Key 
    Lab of Intelligent Technology and Systems, Beijing National Research Center for
    Information Science and Technology,} \\
    {Tsinghua University, Beijing, China.} \\
    \textsuperscript{\rm 2} {Samsung Research China - Beijing (SRC-B), Beijing, China} \\
    \textsuperscript{\rm 3} {\normalsize Fuxi AI Lab, NetEase Inc., Hangzhou, China} \\
    yh.zheng@samsung.com, natnstart@gmail.com, \{zhangrongsheng, huangshilei, maoxiaoxi\}@corp.netease.com, 
	aihuang@tsinghua.edu.cn
}

\begin{document}

\maketitle

\begin{abstract}
Generating stylized responses is essential to build intelligent and engaging dialogue systems. However, this task is far from well-explored due to the difficulties of rendering a particular style in coherent responses, especially when the target style is embedded only in unpaired texts that cannot be directly used to train the dialogue model. This paper proposes a stylized dialogue generation method that can capture stylistic features embedded in unpaired texts. Specifically, our method can produce dialogue responses that are both coherent to the given context and conform to the target style. In this study, an inverse dialogue model is first introduced to predict possible posts for the input responses. Then this inverse model is used to generate stylized pseudo dialogue pairs based on these stylized unpaired texts. Further, these pseudo pairs are employed to train the stylized dialogue model with a joint training process. A style routing approach is proposed to intensify stylistic features in the decoder. Automatic and manual evaluations on two datasets demonstrate that our method outperforms competitive baselines in producing coherent and style-intensive dialogue responses.
\end{abstract}

\section{Introduction}
Building a dialogue agent that can produce stylized and coherent responses has been one of the major challenges in dialogue systems \cite{dinan2019convai2}. Such an agent can yield more vivacious dialogues and deliver more engaging conversations by taking advantage of the linguistic style matching phenomenon~\cite{niederhoffer2002linguistic}, which suggests that people tend to adjust their linguistic style during communication to pursue higher engagement.

\begin{figure}[t]
  \centering
  \includegraphics[width=190px]{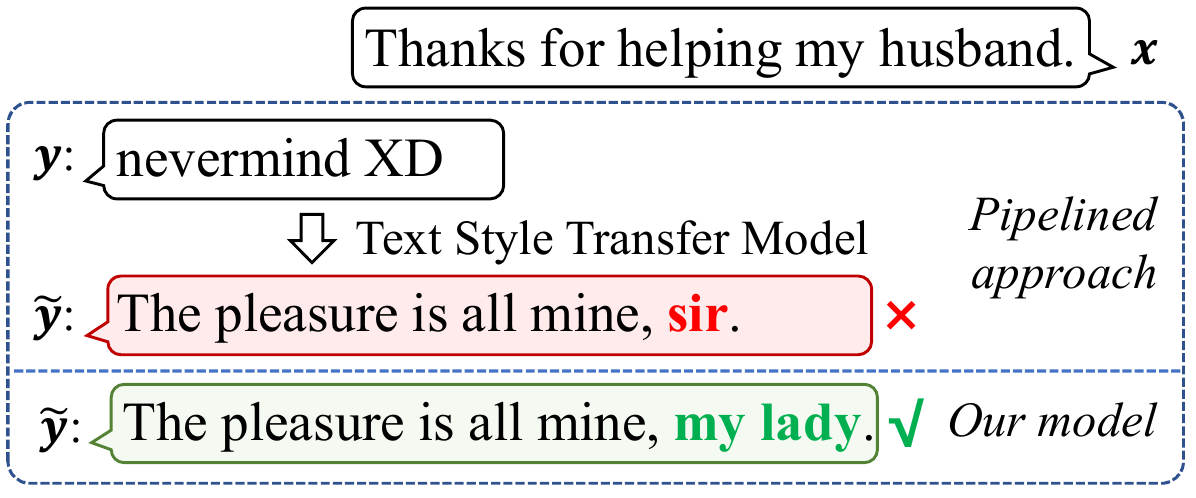}
  \caption{A pipelined approach to produce formal dialogue responses. For a post $x$, a response $y$ is first produced using a dialogue model and then it is transferred to a formal response $\widetilde{y}$ using a text style transfer model. This approach may introduce incoherent contents in $\widetilde{y}$.}
  \label{fig:pipeline_generation}
\end{figure}

Generating stylized dialogue responses has been investigated in various studies, where the definition of styles covers a variety of subtle concepts, such as sentiment~\cite{shen-etal-2017-conditional}, emotion~\cite{zhou2018emotional}, or persona~\cite{Li2016_ACL}. Despite the success, previous studies are generally conducted in a fully supervised setting requiring dialogue pairs in the target style. However, in most cases, the stylistic features we want to capture are embedded in unpaired texts that can not be directly utilized by these supervised models~\cite{gao2019stylefusion}. 

Few studies for dialogue modeling have been proposed to capture the stylistic features embedded in unpaired texts. Specifically, \citet{niu2018polite} and \citet{su2019personalized} employs a style-aware reinforce loss, and \citet{gao2019stylefusion} resorts to a joint continuous latent space. However, despite the reported feasibility, we argue that due to the discrete nature of texts and subtle definition of text styles, it is hard to produce coherent and style-specific responses by relying on sparse reinforce signals or controlling continuous representations.

Note that we can also implement a straightforward stylized dialogue generation pipeline with the help of an unsupervised text style transfer model~\cite{hu2017toward}, which can be trained using stylized unpaired texts. Specifically, for a post $x$, a non-stylized dialogue response $y$ is first generated using a regular dialogue model. Then $y$ is transferred to a stylized response $\widetilde{y}$ using a text style transfer model. However, this approach may hurt the coherence between $x$ and $\widetilde{y}$ since the style transferring process is unaware of $x$ and may introduce inappropriate content. As shown in Figure~\ref{fig:pipeline_generation}, the style transfer model generates a strong stylistic word ``sir'' to emphasize the formality of $\widetilde{y}$. However, this makes $\widetilde{y}$ incoherent with $x$ since $x$ is most likely to be issued by a female.

In this paper, we propose to build a stylized dialogue generation model that can capture stylistic features embedded in a set of unpaired texts $\mathcal{D}_s$. Specifically, to tackle the problem of lacking stylized dialogue pairs, an inverse dialogue model is built to predict posts based on the responses, and a set of stylized pseudo dialogue pairs are constructed by producing pseudo posts for texts in $\mathcal{D}_s$. A stylized dialogue model is then trained using these pseudo pairs, and a joint training process is introduced to enhance the coherency between the post and the resulting responses. Moreover, our dialogue models are parameterized using the Transformer-based encoder-decoder framework and initialized with the pre-trained GPT weights~\cite{radford2018improving}. A style routing approach is devised to fuse a style embedding in each decoder block of the stylized dialogue model to intensify the stylistic features in the decoding process \footnote{See code: \url{https://github.com/silverriver/Stylized_Dialog}}.

We evaluate our method on two datasets with two distinct writing styles: 1) Jinyong novels\footnote{Jinyong is a famous Chinese writer who wrote many Kung Fu novels.} in Chinese, and 2) formality in English writing. Automatic and human evaluations show that our method significantly outperforms competitive baselines with a large margin in generating coherent dialogue responses while rendering stronger stylistic features.

Our contributions can be summarized as:

\textbf{1)} A novel method is proposed to build a stylized dialogue model that can capture stylistic features embedded in unpaired texts. Specifically, an inverse dialogue model is introduced to generate stylized pseudo dialogue pairs, which are further utilized in a joint training process. An effective style routing approach is devised to intensify the stylistic features in the decoder.

\textbf{2)} Automatic and human evaluations on two datasets show that our method outperforms competitive baselines with a large margin in producing stylized and coherent dialogue responses.

\section{Related Work}

\textbf{Stylized dialogue generation} has attracted numerous attention in recent years~\cite{gao2019stylefusion,niu2018polite}. With a rather wide definition of styles, various studies that focus on controllable dialogue generation have been categorized as ``stylized'' dialogue generation, such as generating personalized~\cite{Li2016_ACL,luan-etal-2017-multi,su2019personalized} or emotional~\cite{zhou2018emotional} dialogues. However, these dialogue models' training process usually requires dialogue pairs in the target style, whereas our study aims to capture stylistic features embedded in unpaired texts. 

Moreover, the styles defined in most previous studies are deeply fused with the text contents~\cite{tikhonov2019style}. Enforcing these styles may limit the dialogue model's expressive ability because there are contradictions between certain semantic contents and style categories. For example, it is hard, if not impossible, for a service agent to yield comforting content when enforcing a negative sentiment. Unlike most previous works, our study investigates to model the writing styles that are mostly ``orthogonal'' to the text semantic. The contents we want to deliver will not be constrained by the style we intend to render.

\textbf{Text style transfer} is a related but different task compared to our work. Specifically, these text style transfer models aim to preserve the style-agnostic contents of the input text~\cite{fu2018style}. In contrast, our study aims to produce coherent responses rather than preserve the contents of the posts. Early works on this task focus to disentangle the representation of styles and contents~\cite{hu2017toward,shen2017style,prabhumoye2018style}. However, recent studies argue the effectiveness of such disentanglement~\cite{lample2018multiple} and propose to revise the latent codes using classifiers~\cite{liu2019revision,wang2019controllable}. Some works are also proposed to render the target styles by replacing stylistic words~\cite{wu2019hierarchical,wu2019mask}.

We have also noticed a recent work that considers a contextual constraint in the text style transferring process~\cite{Cheng2020ContextualTS}. However, although being feasible, the training of this model requires style-labeled parallel data. This hinders us from directly employing this model in our study since these parallel data are usually unavailable.

\textbf{Back translation} is a popular approach that has been widely employed in various NLP tasks such as machine translation~\cite{sennrich2015improving}, dialogue data augmentation~\cite{su-etal-2020-diversifying}, text style transfer~\cite{zhang2018style,lample2018multiple,dai2019style}, and even stylized dialogue generation~\cite{wang-etal-2017-steering}. This approach is similar to the inverse dialogue model introduced in our study. However, different from previous approaches that only try to model the one-to-one mapping between the source and target inputs, our inverse dialogue model tries to capture the one-to-many mappings between the responses and posts with the help the proposed joint training process. In our study, the diversity of the generated pseudo posts are enhanced using a sampling approach.

\section{Method}

\subsection{Task Definition}
In this study, we propose to build a stylized dialogue model without utilizing dialogue pairs in the target style. Specifically, our method takes as input two sets of data in the training stage: 1) $M$ unpaired texts $\mathcal{D}_s=\{t_1, ..., t_M \}$ in the writing style $S_1$; 2) $N$ dialogue pairs $\mathcal{D}_p=\{\langle x_1, y_1 \rangle, ...,  \langle x_N, y_N \rangle\}$ with style $S_0$, where $x_i$ and $y_i$ is the post and response, respectively. Our stylized dialogue model aims to generate a response $y$ that is coherent to a given post $x$ while exhibiting a certain style $S_i$ ($i=0, 1$):
\begin{equation}
    y = \argmax_{y'} p({y'}|x, S_i).
\end{equation}

\begin{figure}[tb]
  \centering
  \includegraphics[width=210px]{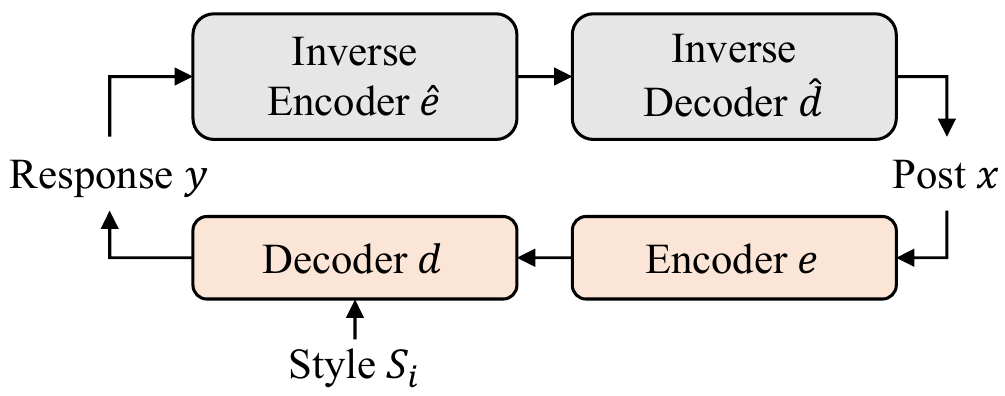}
  \caption{Overall framework.}
  \label{fig:inverse_dialog}
\end{figure}

\subsection{Model Overview}
Our model consists of two mirrored sub-modules (Figure~\ref{fig:inverse_dialog}): (1). A stylized dialogue module (i.e., $e$ and $d$ in Figure~\ref{fig:inverse_dialog}) that can produce a stylized response $y$ based on a given post $x$ and a style label $S_i (i=0,1)$. A style routing approach is devised to incorporate stylistic features in $d$; (2). An inverse dialogue module (i.e., $\hat{e}$ and $\hat{d}$ in Figure~\ref{fig:inverse_dialog}) that aims to produce pseudo posts $x$ based on an input response $y$. Note that the inverse dialogue model is introduced to tackle the problem of lacking dialogue pairs in style $S_1$, i.e., we can regard the texts in $\mathcal{D}_s$ as possible dialogue responses and use the predicted pseudo posts to construct pseudo dialogue pairs in style $S_1$. Therefore, we omit the style label in the inverse decoder $\hat{d}$ to encourage it to focus more on the semantic aspect of the dialogue.

The dialogue modules in our study are parameterized using the Transformer-based encoder and decoder architecture~\cite{Vaswani2017Attention} and are initialized using pretrained GPT~\cite{radford2019language} weights. Further, we also follow previous works~\cite{golovanov-etal-2019-large} to share the weights of the encoder and decoder from the same sub-module to save memories. Particularly, the weights of $e$ and $d$ are shared, and the weights of $\hat{e}$ and $\hat{d}$ are shared.

Moreover, to better capture the one-to-many phenomenon and alleviate the problem of producing trivial posts in the inverse dialogue model, a top-k sampling scheme is employed to sample multiple pseudo posts for each stylized text in $\mathcal{D}_s$. All these sampled posts are utilized in the training process. Further, a joint training process is also introduced to train these two sub-modules iteratively to enhance the coherency of the response.

\begin{figure}[t]
  \centering
  \includegraphics[width=215px]{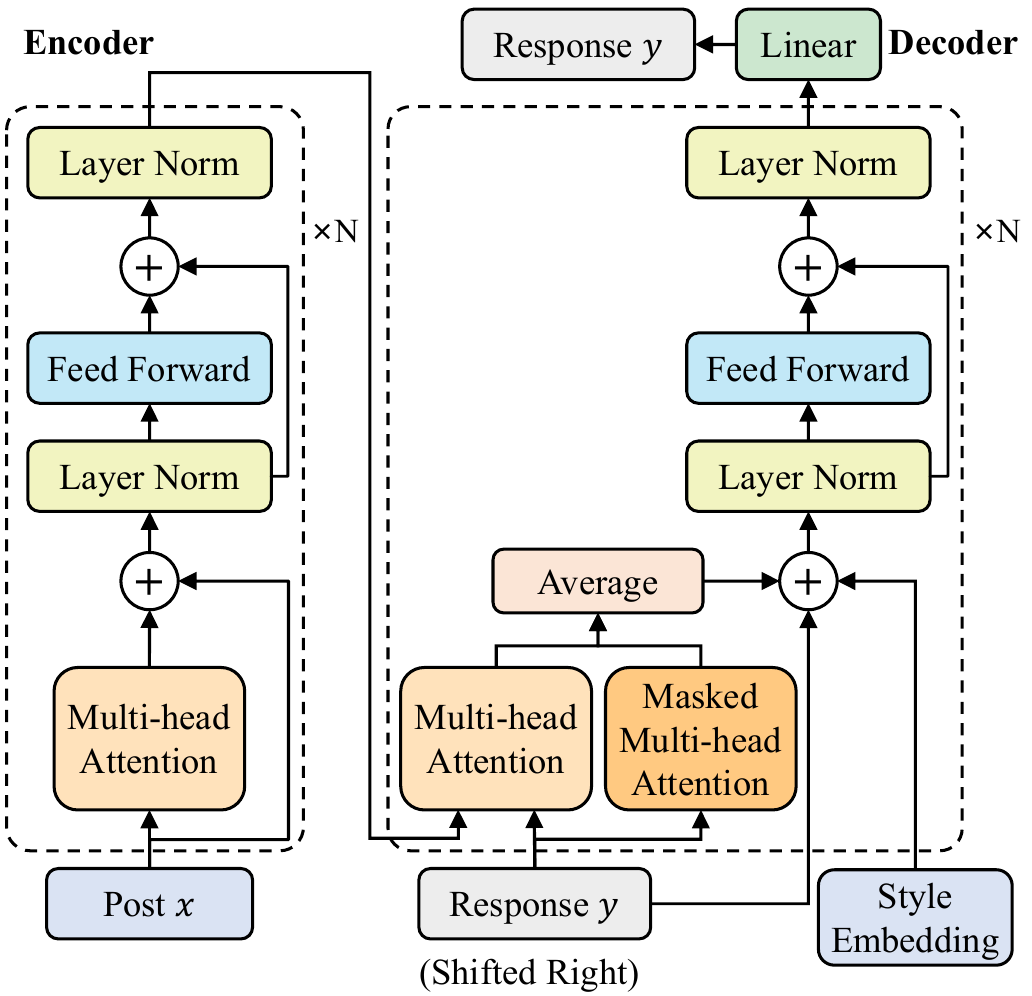}
  \caption{Architecture of the stylized dialogue model.}
  \label{fig:dialog_model}
\end{figure}

\subsection{Style Routing}
There exist various approaches to condition the decoder $d$ on the style label. For example, employing a special style token as the start token~\cite{lample2018multiple}, or adding a style embedding to each word embedding~\cite{zheng2019pre}. However, these approaches only incorporate the style representation in the decoder's input layer, whereas the higher layers are not explicitly affected.

In this study, a style routing approach is devised to enhance existing approaches to stylize $d$ in the stylized dialogue model (see Figure~\ref{fig:dialog_model}). Specifically, in each decoder block, we first fuse the representation of the post $x$ and previously decoded token sequence $y_p$ using the attention routing mechanism~\cite{zheng2019pre}, i.e., two sequences of representations, $\bm{R}_{prev}, \bm{R}_{post} \in R^{l \times h}$, are first calculated:
\begin{align}
\bm{R}_{prev} &= {\rm MMHA}[\bm{e_w}(y_p), \bm{e_w}(y_p), \bm{e_w}(y_p)]\label{eq:post_att},\\
\bm{R}_{post} &= {\rm MHA} [\bm{e_w}(y_p), \bm{e}(x), \bm{e}(x)]\label{eq:prev_att},
\end{align}
where $\bm{e_w}(y_p) \in R^{l \times h}$ denotes the embedding of $y_p$ and it is used as the query in MMHA and MHA, which represent the masked and un-masked multi-head attention operation, respectively. $l$ is the length of $y_p$, and $h$ is the hidden size. $\bm{e_w}(x)$ is the output of the encoder. A sequence of fused representations $\bm{R}_{avg}$ is obtained as:
\begin{align}
    \bm{R}_{avg} = (\bm{R}_{prev} + \bm{R}_{post}) / 2.\label{eq:get_avg}
\end{align}
Then for a given style $S_i$, a style embedding $\bm{e_s}(S_i) \in R^{1 \times h}$ is allocated and $\bm{e_s}(S_i)$ is routed into $\bm{R}_{avg}$ by adding it to each time step of the sequence:
\begin{align}
    \bm{R}_{merge} = \bm{R}_{avg} + \bm{e_s}(S_i).\label{eq:fusion}
\end{align}

Also note that the fusion operation in Eq. \ref{eq:get_avg} and \ref{eq:fusion} is similar to some previous studies that try to incorporate additional contexts in a transformer-based decoder~\cite{golovanov-etal-2019-large}. However, different from these approaches that focus to model sequential contexts, the styles modeled in our study are categorical, and more priority is allocated to the style representation in our model. Moreover, we are the first to use such a style routing approach in the stylized dialogue generation task.

\begin{algorithm}[tb]
\caption{Joint training process} \label{alg:train_process}
\small
\hangindent 3em \textbf{Input}:
$M$ unpaired texts: $\mathcal{D}_s$=$\{t_i\}_{i=1}^M$ in style $S_1$,

\hspace{25px}
$N$ dialogue pairs $\mathcal{D}_p$=$\{\langle x_i,y_i \rangle \}_{i=1}^N$ in style $S_0$.

\textbf{Output}: A stylized dialogue model
\begin{algorithmic}[1] 
\STATE Init the stylized and inverse dialogue model $e$, $d$, $\hat{e}$, $\hat{d}$
\WHILE{not converge}
\STATE Sample $n_d$ dialogue pairs $\mathcal{D}_p^b = \{\langle x_i,y_i \rangle \}_{i=1}^{n_d} \subset \mathcal{D}_p$
\STATE Train $e$ and $d$ by optimizing $\mathcal{L}_{p2r}$ (Eq.~\ref{eq:p2r}) on $\mathcal{D}_p^b$
\STATE Train $\hat{e}$ and $\hat{d}$ by optimizing $\mathcal{L}_{r2p}$ (Eq.~\ref{eq:r2p}) on $\mathcal{D}_p^b$
\IF{Current Step $>$ $N_{f}$}
\STATE $\mathcal{D}_{pp}$ $\leftarrow$ empty set.
\STATE Sample $n_s$ stylized texts $\mathcal{D}_s^b = \{t_i\}_{i=1}^{n_s} \subset \mathcal{D}_s$
\FOR{each $t_i \in \mathcal{D}_s^b$}
\STATE Decode $m$ posts $\{x'_{ij}\}_{j=1}^m$ from $p_{\hat{d}}(x|\bm{\hat{e}}(t_i))$
\STATE $\mathcal{D}_{pp}$ $\leftarrow$ $\mathcal{D}_{pp} \bigcup \{\langle x'_{ij}, t_i\rangle\}_{j=1}^{m}$
\ENDFOR
\STATE Train $e$ and $d$ by optimizing $\mathcal{L}_{inv}$ (Eq.~\ref{eq:inv_loss}) on $\mathcal{D}_{pp}$
\ENDIF
\ENDWHILE
\end{algorithmic}
\end{algorithm}

\subsection{Joint Training}
The training of our model involves the following losses:
1) standard maximum log likelihood losses evaluated on dialogue pairs from $\mathcal{D}_p$:
\begin{align}
\mathcal{L}_{p2r} &= 
\mathop{\mathbb{E}}_{\langle x, y\rangle \sim \mathcal{D}_p} -{\rm log} p_d(y|\bm{e}(x), S_0),\label{eq:p2r} \\
\mathcal{L}_{r2p} &= 
\mathop{\mathbb{E}}_{\langle x, y\rangle \sim \mathcal{D}_p} -{\rm log} p_{\hat{d}}(x|\bm{\hat{e}}(y)).\label{eq:r2p}
\end{align}
The loss $\mathcal{L}_{p2r}$ and $\mathcal{L}_{r2p}$ is used to train the stylized dialogue model and inverse dialogue model, respectively;
2) an inverse dialogue loss evaluated on texts from $\mathcal{D}_s$:
\begin{align}
\mathcal{L}_{inv} =
\mathop{\mathbb{E}}_{\substack{t \sim \mathcal{D}_s,\\ x' \sim p_{\hat{d}} (x| \bm{\hat{e}}(t))}} -{\rm log} p_d(t| \bm{e}(x'), S_1), \label{eq:inv_loss}
\end{align}
in which $x'$ is the pseudo post sampled from the inverse dialogue model.

Note that the gradient back-propagation through the loss $\mathcal{L}_{inv}$ is intractable due to the in-differentiable sampling process in Eq.~\ref{eq:inv_loss}. In this study, we approximate the ideal back-propagation process through $\mathcal{L}_{inv}$ by truncating the gradients associated with the sampling operation. Specifically, when optimizing $\mathcal{L}_{inv}$, the parameters of the inverse dialogue model are fixed, and the stylized dialogue model is trained with pseudo posts $x'$ that are sampled from the inverse dialogue model. Similar approaches have been proven to be effective in other NLP tasks~\cite{lample2018unsupervised,he2020probabilistic}. However, unlike previous works that use the greedy decoding scheme, our study employs the top-k sampling scheme with beam search to produce $x'$ since the mapping between dialogue responses and posts is not unique. The greedy decoding scheme may limit the diversity of the decoded pseudo posts and lead to sub-optimal performance.

\begin{table}[!tb]
\small
\setlength\tabcolsep{3.2pt} 
\centering
\begin{tabular}{lc|cc|cc}
\toprule  
 \multicolumn{2}{c|}{\multirow{2}{*}{Dataset}} & \multicolumn{2}{c}{Train} & \multicolumn{2}{|c}{Test}  \\
                       \cmidrule{3-6}
                      & & $\mathcal{D}_p$ & $\mathcal{D}_s$ & \multicolumn{2}{c}{$\mathcal{D}_t$}  \\
\midrule    
\multirow{2}{*}{WDJN}  & Size & 300.0K   & 95.13K & 2.0K & 2.0K \\
& Style & Weibo & Jinyong & Weibo & Jinyong \\
\midrule
\multirow{2}{*}{TCFC}  & Size & 217.2K   & 500.0K & 0.97K & 0.97K \\
& Style & Informal & Formal & Informal & Formal \\
\bottomrule
\end{tabular}
\caption{Statistics of datasets}
\label{tab:retrival_generation_res}
\end{table}

To facilitate the learning with the above gradient approximation approach, a joint training process is introduced to train the model iteratively. Specifically, in each training iteration, we first update the stylized and inverse dialogue model by optimizing the losses $\mathcal{L}_{p2r}$ and $\mathcal{L}_{r2p}$ using a batch of dialogue pairs sampled in $\mathcal{D}_p$. Further, a batch of stylized sentences $\mathcal{D}_s^b$ are sampled from $\mathcal{D}_s$. For each sentence $t_i \in \mathcal{D}_s^b$, $m$ pseudo posts $x'_{i1}, ..., x'_{im}$ are sampled from the inverse dialogue model, and $m$ pseudo dialogue pairs $\langle x'_{ij}, t_i \rangle, (j=1,...,m)$ in the style $S_1$ are constructed. These pseudo pairs are used to train the stylized dialogue model with the loss $\mathcal{L}_{inv}$. To avoid corrupted pseudo posts at the beginning of the training process, we pre-train the inverse dialogue model on $\mathcal{L}_{r2p}$ for $N_f$ steps before using it to decode pseudo posts. The detailed training process is summarized in Algorithm~\ref{alg:train_process}.

\section{Experiment}
\subsection{Dataset}

Our method is evaluated on two datasets with two distinct styles (see statistics in Table~\ref{tab:retrival_generation_res}).

\textbf{1) WDJN}: We collected 300K \underline{W}eibo \underline{D}ialogues (style $S_0$) as $\mathcal{D}_p$ and sampled 95.1K stylized unpaired texts that are wrapped in quotation marks in \underline{J}inyong's \underline{N}ovels (style $S_1$) as $\mathcal{D}_s$. The texts in $\mathcal{D}_s$ are mostly ``spoken utterances'' that are issued by novel characters.

The testing data of the WDJN dataset $\mathcal{D}_t$ also involves two parts: The first part contains 2.0K additional dialogue pairs collected from Weibo; The second part contains 2.0K dialogue pairs extracted from Jinyong's novel. Specifically, we regard two consecutive spoken utterances that are wrapped in quotation marks as a dialogue pair. We filtered $\mathcal{D}_s$ and $\mathcal{D}_t$ to avoid overlaps.

Also note that to prevent the model from copying stylistic phrases in the post when producing Jinyong style responses in the testing phase, we erased the stylistic features related to Jinyong's writing from the \emph{posts} in these 2K Jinyong style dialogues in $\mathcal{D}_t$ using the back translation approach \cite{zhang2020parallel}. Moreover, all the resulting posts are manually checked and revised to ensure the stylistic features related to style $S_1$ are erased. More details about the WDJN dataset can be found in Appendix A. The WDJN dataset will be released for public use.

\textbf{2) TCFC}~\cite{wu2020dataset}: This dataset focuses on the formality in English writing. We sampled 217.2K informal dialogue pairs (style $S_0$) as $\mathcal{D}_p$ and 500.0K formal texts (style $S_1$) as $\mathcal{D}_s$ from the original dataset, and used the test data in the original dataset as our test set $\mathcal{D}_t$, which contains 1,956 manually-crafted dialogue pairs (978 informal pairs and 978 formal pairs).

\subsection{Implementation Details}
For experiments on the WDJN and TCFC dataset, we used the pre-trained CDial-GPT~\cite{wang2020chinese} and DialoGPT (size 345M)~\cite{zhang2019dialogpt} model to initialize the dialogue modules, respectively. The top-K sampling process in Algorithm~\ref{alg:train_process} employs a $K=20$ and beam size of 4 (WDJN) or 2 (TCFC). The value of $N_f$ is set to 300. The training of our model stops after 10 iteration epochs on $\mathcal{D}_p$ (WDJN) or after 8,000 steps of updates (TCFC). See Appendix B for more details of the reproduction guidance.

\begin{table*}[!t]
\small
\setlength\tabcolsep{3.2pt} 
\centering
\begin{tabular}{L{37pt}lllll|llll||lllll|llll}
\toprule
\multirow{2}{*}{Model} & \multicolumn{9}{c||}{WDJN Dataset} & \multicolumn{9}{c}{TCFC Dataset} \\
\cmidrule{2-19}
 &\multicolumn{2}{c}{BLEU-1,2}  & Dist.   & BERT & SVM & Flu. & Coh. & Style & HAvg. &
  \multicolumn{2}{c}{BLEU-1,2}  & Dist.   & BERT & SVM & Flu. & Coh. & Style & HAvg. \\
\midrule
SLM    & 2.90 & 0.37 & 26.6 & 26.7 & 40.7 & 1.96$^*$ &1.52&0.37&0.79&
         12.6 & 0.99 & 42.5 & 85.6 & 87.2 & 1.90$^*$ &0.89&1.78&1.36 \\
SRL    & 2.53 & 0.33 & 40.4 & 36.2 & 43.2 & 1.83&1.52&0.39&0.82&
         7.83 & 0.70 & 42.7$^*$ & 47.6 & 53.5 & 1.76&0.72&1.25&1.09 \\
SFusion& 3.84 & 0.20 & 33.1 & 8.24 & 19.8 & 1.63&0.69&0.40&0.67&
         5.51 & 0.28 & \B{61.0} & 21.9 & 39.0 & 1.47&0.56&1.17&0.90 \\
S2S+BT & 6.22 & 0.68 & 30.7 & 66.0 & 83.6 & 1.89&1.53$^*$&0.63&1.09&
         12.1 & 1.25 & 42.0 & 86.3 & 86.8 & 1.58&0.72&1.66&1.14 \\
S2S+CT & 11.3 & 0.62 & 32.4 & 72.3 & 76.4 & 0.45&0.19&\B{1.50}&0.38&
         8.05 & 0.64 & 60.9 & 67.7 & 67.8 & 0.37&0.12&0.64&0.24 \\
S2S+PTO& 3.57 & 0.44 & 32.9 & 35.1 & 43.3 & 1.82&1.54$^*$&0.35&0.75&
         9.55 & 0.84 & 34.5 & 28.6 & 50.3 & 0.35&0.26&0.39&0.32 \\

\midrule                                                                  
Ours   & \B{13.6} & \B{1.53} & \B{42.8} & \B{78.3} & \B{89.3} & \B{1.96} & \B{1.60} & 1.16 & \B{1.48} &
         \B{15.1} & \B{1.71} & 43.4 &\B{97.3} & \B{96.1} & \B{1.90} & \B{1.01} & \B{1.89} & \B{1.46} \\
\midrule                                                                  
Human  & \multicolumn{2}{c}{N/A} & 49.3 & 80.1   & 85.4   & 1.93&1.60&1.53&1.67&
         \multicolumn{2}{c}{N/A} & 62.7 & 89.6   & 85.8   & 1.91&1.18&1.83&1.56 \\
\bottomrule
\end{tabular}
\caption{Automatic and manual evaluation results for responses in style $S_1$.
All differences between our model and baselines are significant with $p$-value \textless 0.05 except for the ones marked with *.}
\label{tab:res_s1}
\end{table*}

\begin{table*}[!t]
\small
\setlength\tabcolsep{3.2pt} 
\centering
\begin{tabular}{L{36pt}lllll|llll||lllll|llll}
\toprule
\multirow{2}{*}{Model} & \multicolumn{9}{c||}{WDJN Dataset} & \multicolumn{9}{c}{TCFC Dataset} \\
\cmidrule{2-19}
 &\multicolumn{2}{c}{BLEU-1,2}  & Dist.   & BERT & SVM & Flu. & Coh. & Style & HAvg. &
  \multicolumn{2}{c}{BLEU-1,2}  & Dist.   & BERT & SVM & Flu. & Coh. & Style & HAvg. \\
\midrule
S2S    & 8.50   & 2.42   & 35.1   & 97.0   & 93.0   & 1.96$^*$ & 1.73 & 1.86 & 1.85$^*$ &
         6.92$^*$   & 0.61$^*$   & 54.8   & 70.1$^*$   & 60.9   & 1.82$^*$ & 1.16$^*$ & 1.68$^*$ & 1.50$^*$ \\
SFusion& 8.65   & 0.82   & 35.3   & \B{99.9}   & \B{99.2}   & 1.41 & 0.74 & 1.92$^*$ & 1.16 &
         4.61   & 0.22   & \B{62.8}   & \B{70.3}   & \B{61.1}   & 1.57 & 0.76 & \B{1.77}$^*$ & 1.19 \\
\midrule                                                                  
Ours   & \B{11.6} & \B{2.93} & \B{39.0} & 93.5   & 89.2   & \B{1.97} & \B{1.85} & \B{1.93} & \B{1.92} &
         \B{6.96} & \B{0.67} & 49.4     & 69.4   & 59.2  & \B{1.85} & \B{1.16} & 1.70 & \B{1.51} \\
\midrule
Human  & \multicolumn{2}{c}{N/A} & 56.4 & 97.9   & 94.4   & 1.89 & 1.86 & 1.98 & 1.91 &
         \multicolumn{2}{c}{N/A} & 72.6 & 72.0   & 72.1   & 1.76 & 1.19 & 1.76 & 1.52 \\
\bottomrule
\end{tabular}
\caption{Automatic and manual evaluation results for responses in style $S_0$.
All differences between our model and baselines are significant with $p$-value \textless 0.05 except for the ones marked with *.}
\label{tab:res_s0}
\end{table*}

\subsection{Baselines}
We choose two groups of baselines:

The first group contains dialogue models with different style modeling scheme: \textbf{1) \emph{S2S}} \cite{golovanov-etal-2019-large}: a strong Transformer-based dialogue model that is only trained on $\mathcal{D}_p$. This baseline can only produce responses in style $S_0$; \textbf{2) \emph{SLM}}: the ``Fusion'' model proposed by \citet{niu2018polite}, in which an independent stylized language model is trained on $\mathcal{D}_s$, and the distributions decoded from the \emph{S2S} baseline and the stylized LM are fused when producing responses in style $S_1$; \textbf{3) \emph{SRL}}: the ``RL'' model proposed by \citet{niu2018polite}, in which a reinforce signal produced by a style classifier is used to enforce the style $S_1$; \textbf{4) \emph{SFusion}}~\cite{gao2019stylefusion}: A fused latent space is built using a multi-task training scheme. Specifically, for each post, six responses are sampled, and two classifiers are used to rank these responses for the styles.

The second group of baselines are built using the pipelined approach, i.e., different unsupervised text style transfer models are trained on texts from $\mathcal{D}_s$ and $\mathcal{D}_p$, and responses produced by the \emph{S2S} baseline (in style $S_0$) are transferred to exhibit the target style $S_1$ using these models:
\textbf{5) \emph{S2S+BT}}: a back-translation-based text style transfer model~\cite{he2020probabilistic}; 
\textbf{6) \emph{S2S+CT}}: a model that tries to entangle the latent code for styles and contents~\cite{wang2019controllable}; 
\textbf{7) \emph{S2S+PTO}}: a model that renders the target style by replacing stylistic words~\cite{wu2019hierarchical}.

Note that for baselines 2, 3 and 5-7, the responses generated by the S2S baseline are used as their responses for the $S_0$ style since they can only produce responses in $S_1$ once trained. Moreover, we implemented baselines 1-3 using the same architecture and hyper-parameters as our model for fair comparisons. For baselines 4-7, we used the official codes released by the authors. Note that it is non-trivial to utilize the pre-trained GPT model in the baseline \emph{SFusion} since it handles fixed-length latent codes.

\subsection{Automatic Evaluation}

\textbf{Metrics:}
We first used automatic metrics to evaluate the response quality of our model: 1). \textbf{\emph{BLEU}}~\cite{papineni-etal-2002-bleu} was used to measure n-gram (n=1, 2) overlap between the generated responses and the reference responses; 2). \emph{Distinct} (\textbf{Dist.})~\cite{li2015diversity} measures the proportion of unique n-grams in the generated responses (n=2).

To evaluate the style intensity of each model, we first trained two text style classifiers (i.e., \textbf{\emph{BERT}} \cite{devlin2018bert} and \textbf{\emph{SVM}}) and then calculated the style intensity score as the portion of generated responses that conform to the target style based on these classifiers. In our study, texts from $\mathcal{D}_p$ and $\mathcal{D}_s$ were used to train classifiers for the WDJN experiments, and the GYAFC dataset \cite{rao-tetreault-2018-dear} was used to train classifiers for the TCFC experiments. The accuracy of the BERT and SVM classifier on the holdout test set was 98.52\% and 94.20\% respectively for the WDJN experiments, and 93.98\% and 89.57\% respectively for the TCFC experiments (see Appendix C for more details). 

\begin{table*}[!t]
\small
\setlength\tabcolsep{2.8pt} 
\centering
\begin{tabular}{L{40pt}lllll|llll||lllll|llll}
\toprule
\multirow{2}{*}{Model} & \multicolumn{9}{c||}{WDJN Dataset} & \multicolumn{9}{c}{TCFC Dataset} \\
\cmidrule{2-19}
 &\multicolumn{2}{c}{BLEU-1,2} & Dist. & BERT & SVM & Flu. & Coh. & Style & HAvg. & \multicolumn{2}{c}{BLEU-1,2}   & Dist.   & BERT & SVM & Flu. & Coh. & Style & HAvg. \\
\midrule 
Ours       & \B{12.6} & \B{2.23} & \B{40.9} & 85.9 & 89.3 & \B{1.96} & \B{1.69} & 1.54 & \B{1.71} &
             \B{11.0} & \B{1.19} & 46.4 & 83.3 & 77.7 & \B{1.87} & \B{1.08} & 1.79 & 1.49 \\ 
\midrule 
w/o Rout.  & 12.5$^*$ & 2.19$^*$ & 40.3$^*$ & 84.0 & 87.1 & 1.51 & 1.25 & 1.48 & 1.41 &
             \B{11.0}$^*$ & 1.13$^*$ & \B{46.8}$^*$ & 82.8$^*$ & 77.0 & 1.86$^*$ & 1.03 & 1.78$^*$ & 1.45 \\ 
w/o JointT & 7.57 & 1.71 & 30.1 & \B{91.3} & \B{94.5} & 1.51 & 1.23 & 1.54$^*$ & 1.41 &
             9.84 & 0.97 & 46.2$^*$ & 83.9$^*$ & 78.7 & \B{1.87}$^*$ & 0.95 & \B{1.81}$^*$ & 1.40 \\ 
w/o Samp.  & 9.79 & 1.62 & 39.7 & 90.2 & 92.8 & 1.26 & 1.07 & 1.58$^*$ & 1.27 &
             10.7$^*$ & 1.18$^*$ & 46.7$^*$ & 83.4$^*$ & 78.3 & \B{1.87}$^*$ & 0.99 & \B{1.81}$^*$ & 1.43 \\ 
w/o PreT   & 10.9 & 1.43 & 16.9 & 91.2 & 91.9 & 1.46 & 0.90 & \B{1.60}$^*$ & 1.24 &
             10.1$^*$ & 0.74 & 32.5 & \B{84.8} & \B{78.8} & 1.86$^*$ & 0.65 & 1.80$^*$ & 1.15 \\ 
\bottomrule
\end{tabular}
\caption{Automatic and manual evaluation results of ablation models for responses in style $S_0$ and $S_1$.
All differences between our model and ablation models are significant with $p$-value \textless 0.05 except for the ones marked with *.}
\label{tab:ablation_res}
\end{table*}

\begin{figure}[!bp]
  \centering
  \includegraphics[width=210px]{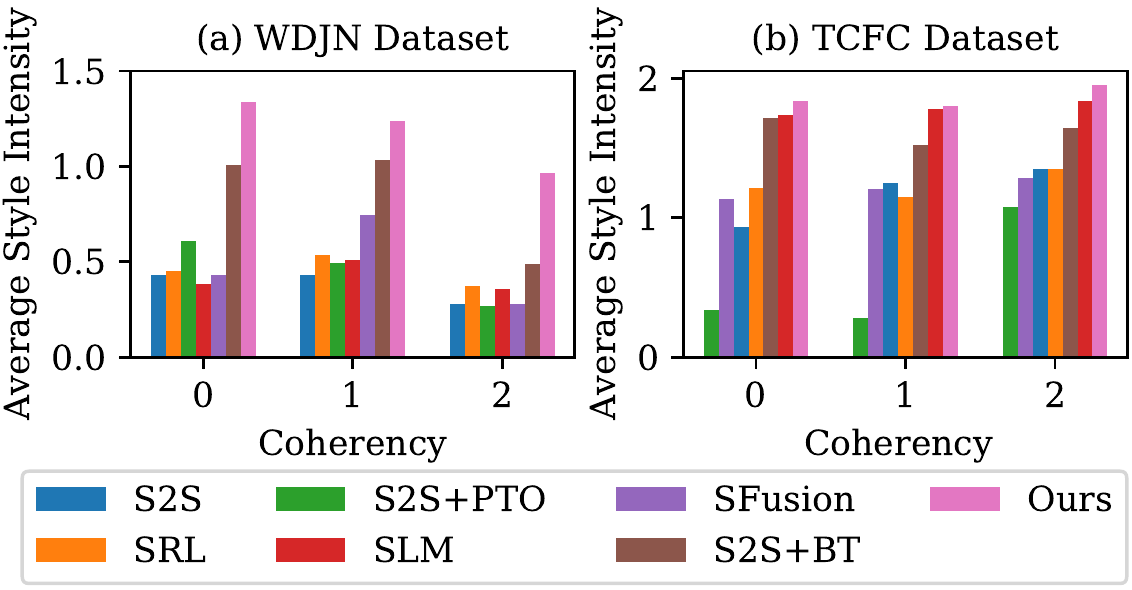}
  \caption{Averaged style intensity scores for responses with different coherency scores.}
  \label{fig:style_with_coh}
\end{figure}

\textbf{Results:}
We separately evaluated the responses in style $S_1$ (Table \ref{tab:res_s1}) and $S_0$ (Table \ref{tab:res_s0}). Note that the baseline \emph{S2S} is not included in Table \ref{tab:res_s1} since it can not produce responses in style $S_1$. Similarly, only the baselines \emph{S2S} and \emph{SFusion} are contained in Table \ref{tab:res_s0}. Significance tests are performed between the results of our model and all the baselines using the t-test with bootstrap resampling~\cite{koehn-2004-statistical}.

As can be seen from the automatic results, our method outperforms all the baselines with large margins when generating dialogue responses in style $S_1$ (Table \ref{tab:res_s1}), and achieves competitive performance when producing responses in style $S_0$ (Table \ref{tab:res_s0}). This indicates that our model can produce high-quality responses that are both coherent with the given context and consistent to the target style. We can further observe that:

1). The pipelined approaches achieve lower BLEU scores comparing to our method. This verifies our claim that the response coherency is affected by the style transferring process. Similar results are also observed in manual evaluation. Also note that some non-pipelined baselines achieve lower BLEU scores comparing to the pipelined baselines, e.g. SRL and SFusion on the TCFC dataset. This indicates that it is hard to capture the stylistic features by modifying the latent spaces.

2). The high diversity (i.e., \emph{Dist.} scores) of the baselines on the TCFC dataset come along with a dramatic decrease on the BLEU scores. This is because these baselines overfit to the diverse colloquial phrases in the informal responses and fail to render responses in style $S_1$, which are more formal and less diverse.

Also note that the style intensity scores for human-generated responses (last row in Table \ref{tab:res_s1} and \ref{tab:res_s0}) do not match the accuracy of our style classifiers. This is because these classifiers' train data involve non-conversational texts, which leads to mismatches when testing using conversational responses. To alleviate this mismatch, we performed manual evaluations to concrete our analysis.

\begin{figure*}[ht]
  \centering
  \includegraphics[width=470px]{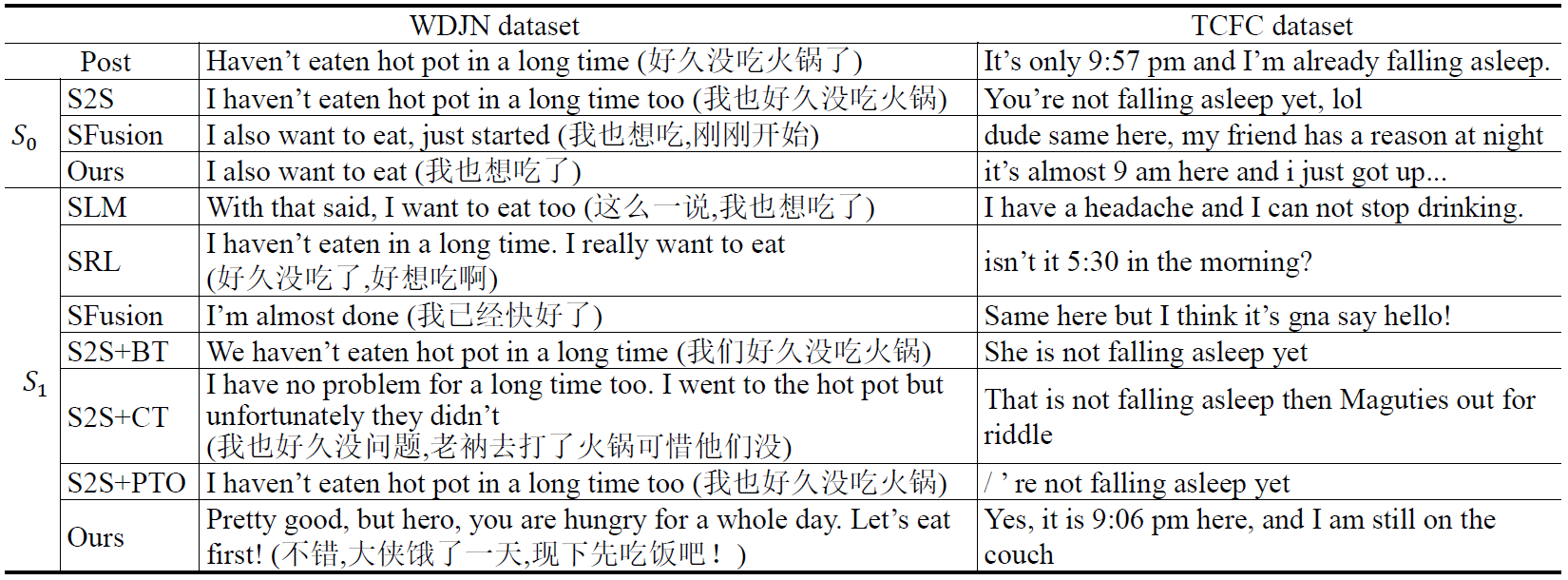}
  \caption{Example responses produced by our model and the baselines on the TCFC and WDJN datasets.}
  \label{tab:example_pair}
\end{figure*}

\begin{figure}[!tbp]
  \centering
  \includegraphics[width=230px]{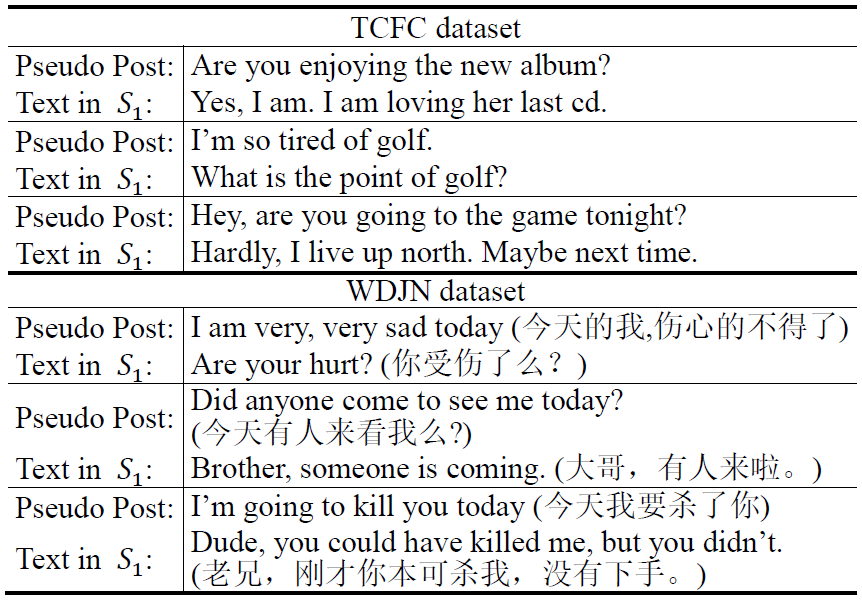}
  \caption{Example pseudo pairs generated by the inverse dialogue model in the training process.}
  \label{tab:pseudo_pair}
\end{figure}

\subsection{Manual Evaluation}

\textbf{Metrics:}
For a given post, dialogue responses with different styles were generated using our model and all the baselines. Three annotators were recruited from the crowd-sourcing platform to evaluate these responses from three aspects: 1) \emph{Fluency} (\textbf{Flu.}): whether the response is fluent and free from grammar errors; 2) \emph{Coherency} (\textbf{Coh.}): whether the response is coherent with the dialogue context; 3) \emph{Style Intensity} (\textbf{Style}): whether the response conforms to the given style. Each metric is rated among \{0, 1, 2\}, in which 0 means worst and 2 best. Moreover, the \emph{Harmonic Average} (i.e, \textbf{HAvg.}) of above measures is also reported.

\textbf{Results:}
We sampled 300 posts from $\mathcal{D}_t$ for each of these two datasets. Fleiss's kappa $\kappa$~\cite{randolph2005free} was used to measure the annotation agreement between annotators. Specifically, for \emph{Flu.}, \emph{Coh.}, and \emph{Style}, the $\kappa$ value was 0.69, 0.50, 0.86, respectively on the WDJN dataset (indicating substantial, moderate, and substantial agreement), and 0.44, 0.31, 0.42, respectively on the TCFC dataset (indicating moderate, fair, and moderate agreement). 

As shown in Table~\ref{tab:res_s1}, our model surpasses all the baselines significantly on style intensity (except for S2S+CT on the WDJN dataset, which comes with dramatic decreases on the fluency and coherency scores) when producing responses in style $S_1$, and it achieves competitive or higher fluency and coherency scores. This verifies the superiority of our method in producing coherent and style intensified dialogue responses. Moreover, results in table~\ref{tab:res_s0} also show that our model achieved competitive performance when generating responses in style $S_0$.

We can also observe from Table \ref{tab:res_s1} and \ref{tab:res_s0} that:

1). There are trade-offs between the coherency and style intensity when generating stylized dialogue responses on the WDJN dataset, i.e., the high style intensity usually comes at the cost of a low coherency. For example, the S2S+CT achieves the best style intensity score on WDJN (1.50) when producing responses in style $S_1$, but it obtains the worst coherency (0.19) score. This phenomenon is also observed in various previous studies~\cite{niu2018polite,zheng2019pre}. Nevertheless, our model achieves competitive coherency while producing style-intensive responses.

2). The distribution mismatch between texts in $\mathcal{D}_s$ and $\mathcal{D}_p$ may bring an adverse impact on the response quality in style $S_1$. For example, on the TCFC dataset, the coherency score of our model drops from 1.18 (human performance) to 1.01 in Table \ref{tab:res_s1}. This is because that the data in $\mathcal{D}_s$ of TCFC originate from the QA texts that are usually not in a conversation form. In contrast, our model obtains the same coherency score with human responses on WDJN since the data in $\mathcal{D}_s$ of WDJN are mostly unpaired conversational texts.

3). The baselines SFusion, SRL, and S2S+CT generally yield low \emph{HAvg.} scores on both datasets. This verifies our claim that it is hard to generate stylized and coherent responses relying on the sparse reinforce signals (i.e., SRL) or continuous latent codes (i.e., SFusion and S2S+CT).

Our method's superiority to generate stylized dialogue responses is further demonstrated by analyzing the style intensity scores of responses with different coherency levels. Specifically, all the annotated responses in style $S_1$ were collected and categorized into three groups based on the coherency scores (i.e., 0, 1, or 2) they received. The averaged style intensity score for each group was calculated and shown in Figure~\ref{fig:style_with_coh}. It can be seen that our model achieves the highest style intensity scores in all coherency groups. This further demonstrates that the responses produced by our method are more style-intensive than those by the baselines. Note that the baseline S2S+CT is not included in Figure~\ref{fig:style_with_coh} because its fluency score is extremely low. There is no point in comparing its coherency to other baselines.

Also note that the results in Figure~\ref{fig:style_with_coh} very much by the dataset. It is easier to capture the stylistic features in the TCFC dataset. This interesting phenomenon may dues to the fact that there are larger gaps between texts in style $S_0$ and $S_1$ in the dataset WDJN. Specifically, style $S_0$ texts in WDJN originate from the web corpus (i.e., Weibo) and style $S_1$ texts originate from Kung-fu novels written in 1960s. Such gaps in TCFC is much smaller since style $S_0$ and $S_1$ texts in TCFC all originate from the web corpus.
This also explains why the the trade-offs between the coherency and style intensity in TCFC dataset is not that significant. For example, it is hard, if not impossible, to describe some modern events (e.g. ipod or 5G networks) using phases in 1960s' Kung-fu novels.

\subsection{Ablation Study}

Ablation studies were performed to verify the effect of each component in our method. Specifically, the following variants were tested: 
1) without the style routing approach (\textbf{w/o Rout.}), i.e., the style embedding is not incorporated in each decoder block as in Eq.\ref{eq:fusion}. The decoder $d$ is stylized by employing stylized start token and adding a style embedding to each word embedding;
2) Without the joint training process (\textbf{w/o JointT}), i.e., an inverse dialogue model is first trained and fixed, and then a fixed set of pseudo pairs are generated and used to train the stylized dialogue model. Note that the same amount of pseudo pairs were used to optimize the loss $\mathcal{L}_{inv}$ in this variant as it is used in Algorithm~\ref{alg:train_process};
3) Without using the top-K sampling scheme when producing pseudo posts (\textbf{w/o Samp.}), i.e., pseudo pairs are decoded greedily;
4) Without using the pre-trained GPT weights (\textbf{w/o PreT}). 

As shown in Table~\ref{tab:ablation_res}, our model achieves the highest \emph{BLEU} and \emph{Coh.} scores among all the ablation models. We can further observe that: 1) Almost all our variants surpass the baselines with a large margin on the style intensity score. This verifies the feasibility of our framework in capturing stylistic features; 2) Removing the joint training process (\textbf{w/o JointT}) or the top-K sampling scheme (\textbf{w/o Samp.}) makes the dialogue models over-fit to render more stylistic features while failing to achieve high \emph{BLEU} and \emph{Coh.} scores. However, we argue that since our stylized decoder is already strong in capturing stylistic features, it is critical to utilize the proposed joint training and top-K sampling scheme to improve the response coherency; 3) The pre-training approach significantly improves the diversity and coherency of the generated responses.

\subsection{Case Study}

Figure~\ref{tab:example_pair} shows some dialogue responses generated by our model and the baselines on the two datasets. We can observe that the models that directly manipulate the continuous latent space (i.e., SFusion and S2S+CT) yield non-fluent responses. 
This is because it hard to build a smooth latent space for discrete texts. 
Moreover, on the TCFC dataset, the baseline SFusion in style S0, and baselines SLM, SFusion, S2S+BT, S2S+CT, S2S+PTO in style S1 fail to produce coherent responses to the post.
Further, pipelined approaches either fail to convert the inputs to the target style (i.e., S2S+PTO on the WDJN dataset) or hurt the coherency between the response and the post (i.e., S2S+BT, S2S+CT, and S2S+PTO on the TCFC dataset).

In addition, we sampled some of these pseudo pairs generated by the inverse dialogue model in the training phase (Figure~\ref{tab:pseudo_pair}). It can be seen that these pseudo pairs are generally of high quality both in fluency and coherency.

\section{Conclusion}
This paper presents a stylized dialogue generation method that can produce coherent and style-intensive responses by utilizing stylized unpaired texts. An inverse dialogue model is introduced in our method to produce stylized pseudo dialogue pairs, which are used in a joint training process to train the stylized dialogue model. Further, a style routing approach is introduced to intensify stylistic features in the decoding process. We demonstrate our method on two datasets with two different styles: Chinese Jinyong novels and English formality writing. The automatic and manual evaluation shows that our method outperforms competitive baselines in producing coherent and style-intensive responses. As future works, we will extend this method to other stylized text generation tasks.

\section{Acknowledgement}
This work was jointly supported by the Major Project of the New Generation of Artificial Intelligence(No. 2018AAA0102900),
and NSFC projects (Key project with No. 61936010 and regular project with No. 61876096). 
This work was also supported by the Guoqiang Institute of Tsinghua University,
with Grant No. 2019GQG1. We thank THUNUS NExT Joint-Lab for the support.

\bibliography{ref.bib}

\end{document}